\documentclass[runningheads]{llncs}
\usepackage[T1]{fontenc}
\usepackage{graphicx}
\usepackage{algorithm}
\usepackage{algpseudocode}
\usepackage{amsmath}
\usepackage{tabularx} 
\usepackage{tikz}
\usepackage{verbatim}
\usepackage{array,tabularx}
\begin{document}
\title{Who Sees What? Structured Thought-Action Sequences for Epistemic Reasoning in LLMs}
\titlerunning{Who Sees What?}
\author{Luca Annese \inst{1}, Sabrina Patania \inst{1}, Silvia Serino \inst{1},Tom Foulsham\inst{3},  Silvia Rossi \inst{2}, Azzurra Ruggeri \inst{4} and Dimitri Ognibene\inst{1,3}}
\institute{University of Milan-Bicocca, Milan, Italy\\ 
\email{\{luca.annese1,sabrina.patania,silvia.serino,dimitri.ognibene\}@unimb.it}  \\
\and University of Essex, Colchester, UK
\email{foulsham@essex.ac.uk}\and University of Naples Federico II, Naples, Italy
\email{silvia.rossi@unina.it}
\and TUM School of Social Sciences and Technology, Munich, Germany\email{}}
\authorrunning{L. Annese et al.}

%
\maketitle              % typeset the header of the contribution
\begin{abstract}
Recent advances in large language models (LLMs) and reasoning frameworks have opened new possibilities for improving the  perspective -taking capabilities of autonomous agents. 
However, tasks that involve active perception, collaborative reasoning, and perspective taking (understanding what another agent can see or knows) pose persistent challenges for current LLM-based systems.
This study investigates the potential of structured examples derived from transformed solution graphs generated by the Fast Downward planner to improve the performance of LLM-based agents within a ReAct framework. We propose a structured solution-processing pipeline that generates three distinct categories of examples: optimal goal paths (G-type), informative node paths (E-type), and step-by-step optimal decision sequences contrasting alternative actions (L-type). These solutions are further converted into ``thought-action'' examples by prompting an LLM to explicitly articulate the reasoning behind each decision.
While L-type examples slightly reduce clarification requests and overall action steps, they do not yield consistent improvements. Agents are successful in tasks requiring basic attentional filtering but struggle in scenarios that required mentalising about occluded spaces or weighing the costs of epistemic actions. These findings suggest that structured examples alone are insufficient for robust perspective-taking, underscoring the need for explicit belief tracking, cost modelling, and richer environments to enable socially grounded collaboration in LLM-based agents.
 %Our evaluations in environments with increasing demands on perspective-taking and partial observability revealed modest performance improvements exclusively when incorporating L-type examples. Although agents successfully retrieved correct objects, performance notably declined in scenarios requiring inference about objects that were initially not visible. This underscores the complexity of multimodal collaboration in partially known environments, where agents must balance sophisticated social reasoning with estimations of epistemic and empirical costs and gains—capabilities not fully realizable with current LLM-based reasoning frameworks, even when enhanced by structured examples. Consequently, our findings emphasize the need for further research into integrating these advanced cognitive skills into foundational robotics models to achieve effective multi-agent collaboration.
\end{abstract}
\section{Introduction}
Effective interaction in multi-agent systems, especially those involving human-Artificial intelligence (AI) collaboration, requires more than basic task execution. It demands the capacity for perspective-taking: the ability to model what others can see, know, or intend. This includes reasoning about both physical viewpoints (e.g., what another agent can perceive in the environment) and epistemic states (e.g., what they know or believe). Perspective-taking is essential for generating contextually appropriate responses and adapting to ambiguity in real-world scenarios.

In this work, we explore how perspective-taking can be operationalized within the ReAct (Reason+Act) framework \cite{yao2023react}, which interleaves natural language reasoning with environment-grounded actions. Unlike static perception-action pipelines, ReAct enables agents to reason explicitly before acting, providing an ideal foundation for handling the interpretive demands of perspective-taking tasks.

We focus on a modified version of the Director Task \cite{keysar2000taking}, adapted to a partially observable environment inspired by \cite{sarthou2021director}. In our setting, already explored in \cite{patania2025perspact} and  \cite{patania2025growing}, a Director issues instructions to a Matcher agent, which must retrieve a target object. Crucially, the environment contains occlusions and hidden containers that affect both agents perception, meaning that the Matcher may have to actively perceive \cite{ognibene2013contextual,ognibene2019proactive} to infer what the Director sees and does not see, to resolve ambiguity. This setup simulates real-world collaborative scenarios where agents have asymmetric and limited access to information.

To enable grounded perspective-taking, we introduce a novel training method that builds on search-based reasoning from the Fast Downward planner. 
%Starting  from complete reasoning trees that encode  possible thought-action trajectories in related conditions The agent is provided with three types of examples constructed :
%Before agent execution,
We construct complete reasoning trees that encode possible thought-action trajectories in similar conditions. 
From these trees, we extract three types of example sequences: 
\begin{itemize} 
\item G-type: optimal sequence from initial state to task success; 
\item E-type: all paths that reach informative states, where the agent gains new knowledge through sensing, movement, or object interaction; 
\item L-type: locally optimal decisions at each state, simulating reactive step-wise decision-making. \end{itemize} 
Each sequence is transformed into a chain of thought-action pairs using a LLM, enabling agents to learn not just what to do, but why, increasing both interpretability and transferability.

We evaluate our method across seven environments of increasing complexity, varying both the spatial layout and the ambiguity of instructions to intensify the perspective-taking demands. For each trial, the agent is shown examples from six environments and evaluated in the held-out seventh, testing generalization across diverse settings. Tasks ranged from fully disambiguated instructions to scenarios requiring the agent to infer what the Director sees or does not see.
Empirical results show that while G-type and E-type examples support task efficiency and exploration, only L-type examples slightly improved agent behaviour, specifically by reducing excessive clarification requests and promoting more grounded decision-making.

To interpret these results, we introduce a functional characterisation of the cognitive demands associated with different task conditions. These demands (ranging from simple common-ground attentional filtering \cite{keysar1997unconfounding}, to mentalising about hidden content \cite{bianco2022robot}, and metacognitive evaluations of epistemic cost and utility \cite{keysar2003limits}) emerge in increasingly complex environments and help explain when LLM agents succeed and where they systematically fail.

Ultimately, while embedding cognitively structured examples into the ReAct loop scaffolds some forms of perspective-taking—particularly those grounded in reactive reasoning and observable cues—it may not suffice for enabling higher-order inference, imaginative simulation, or consistent collaborative strategies that unfold over multiple steps. Such strategies require balancing social reasoning, physical exploration, and interlocutor modelling, while accounting for both information gain and the cost or risk of failure.
While recent work suggests that LLMs exhibit emerging capabilities for social reasoning and information gathering \cite{patania2024large,amirizaniani2024can}, our findings underscore the need for their integration with socially-aware, active information-seeking strategies \cite{ognibene2019proactive,ognibene2013contextual}, in order to support robust collaboration in open-ended, multi-agent environments \cite{bianco2019functional,doi:10.1089/cyber.2024.0536}.

\section{Related work}
In recent years, there has been growing interest in the application of large language models (LLMs) and multimodal foundation models in robotics and collaborative systems for high-level reasoning, perception, and decision-making \cite{ognibene2025scoopframeworkproactivecollaboration}. These models are pre-trained on vast amounts of internet-scale data and exhibit impressive generalization capabilities \cite{brown2020language}, enabling robots to handle a wide range of open-ended scenarios. Models such as SayCan \cite{ahn2022can} and Inner Monologue \cite{huang2022inner} demonstrate how LLMs can break down abstract goals into practical steps by combining high-level reasoning with grounded robotic actions.

A core component of effective multi-agent interaction is perspective-taking, namely, the ability to represent a situation from an alternate viewpoint \cite{grice1975logic}. This includes visual perspective-taking, distinguished between Level-1 (inferring what others can see) and Level-2 (inferring how things appear to others), and spatial perspective-taking, which involves representing relative spatial relations through egocentric or allocentric reference frames \cite{flavell1981young}. Visual perspective-taking, particularly Level-2, has been closely linked to theory of mind (ToM), as both require agents to maintain decoupled mental representations. In frameworks like ReAct \cite{yao2023react}, perspective-taking is framed as a dynamic reasoning process that accompanies acting, enabling agents to update their knowledge in real time.

Efforts to enhance perspective-taking in LLMs have largely focused on language based evaluations. Studies using false-belief tasks indicate that while older models (e.g., GPT-2, early GPT-3) struggle with ToM tasks, more recent systems (e.g., GPT-4) display emerging but unstable capabilities \cite{kosinski2024evaluating}. Techniques like the SimToM prompting framework \cite{wilf2024think} explicitly instruct models to simulate other agents’ perspectives, reducing the intrusion of background omniscience. In the visual domain, datasets such as Isle-Bricks and Isle-Dots \cite{goral2024seeing} reveal that while many vision-language models (VLMs) can detect objects in a scene, they often fail at reasoning about what is visible from an observer’s viewpoint. Advanced models like GPT-4V perform well on Level-1 tasks but show notable drops on Level-2 challenges involving viewpoint rotation and mental transformation \cite{leonard2024failures}. Beyond static perception, benchmarks such as ActiView \cite{wang2024actiview} introduce active visual exploration, requiring models to shift or zoom their viewpoint to gather relevant information—tasks that remain difficult for current models.

Complementary to these efforts, recent research has begun to explore the synergy between LLMs and classical symbolic planning systems to enhance structured reasoning. Hybrid models like LLM+P \cite{liu2023llmpempoweringlargelanguage} use planners to generate plans from formal representations (e.g., PDDL), then convert them into natural language for execution by LLMs. Other works, such as PSALM \cite{zhu2024languagemodelsinferaction}, investigate using LLMs to synthesize or refine planning domains, while others employ LLMs to verbalize or critique symbolic plans \cite{huang2022inner}. These approaches leverage the complementary strengths of symbolic methods (e.g., correctness, structure) and language models (e.g., flexibility, generalization).

Building on these developments, our work explores how symbolic planning can serve as a source of structured, cognitively meaningful training examples for LLMs. This method bridges symbolic and neural paradigms by using planning structures to ground language-based cognitive traces, offering a new way to study and enhance perspective-taking and decision-making in LLM-based agents.

\section{Method}
\begin{figure}[!h]
    \centering
    \includegraphics[width=1\textwidth]{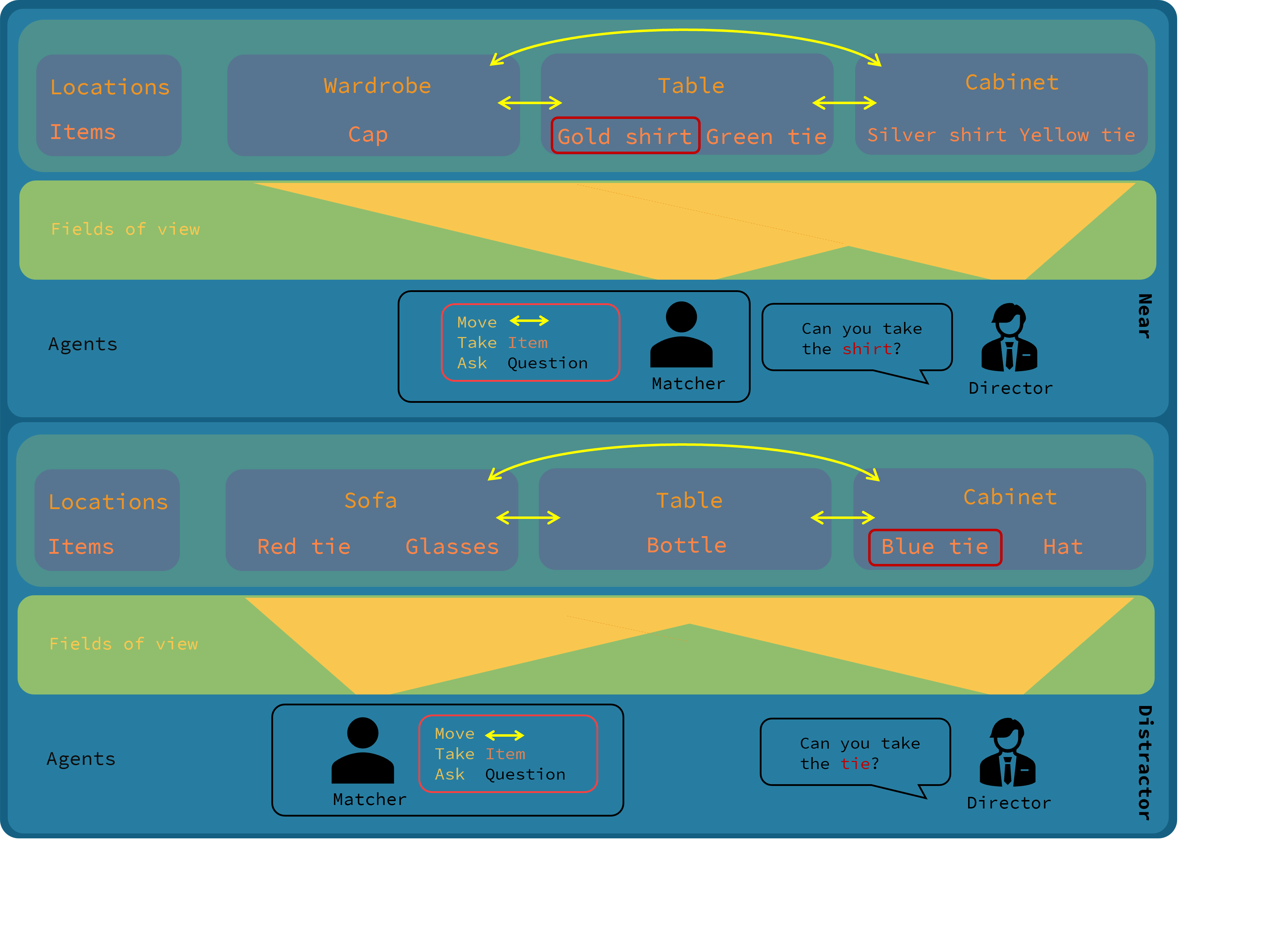}
    \caption{Schematic view showing two examples of the experimental environment. In each case the top row shows the three locations and their item contents, and the yellow arrows indicate possible movements from the Matcher. The target is outlined in red.
    Each agent can see only the location in front and immediately adjacent (yellow shading). The Matcher can move between locations, take an item from the location directly in front, or ask the Director a question. The Director can only answer from their fixed position. In the Near condition (top panel), there is ambiguity because two candidate objects fit the Director's request (Gold shirt and Silver shirt). In the Distractor condition (bottom panel), the Matcher can see an item (Red tie) that matches the request, yet is not the correct target. }
    \label{fig:env}
\end{figure}
\subsection{Simulated Task Environment}
To explore perspective-taking in goal-directed interaction, we developed a simulated household-like environment using the Planning Domain Definition Language (PDDL). The environment represents a shared space between two agents: a Director, who knows the identity and location of the target object, and a Matcher, who must retrieve the object based on limited perceptual cues and dialogue. The space includes multiple locations such as a desk, shelf, and drawers, with some objects hidden inside containers that can be opened or closed.

The PDDL domain models both spatial structure and perceptual asymmetry. Each agent can perceive the contents of its own location, as well as any adjacent locations, simulating a partially shared field of view. This shared access supports basic grounding, while still requiring inference about what the other agent can or cannot see.

% \begin{table}[!h]
% \centering
% \small
% \renewcommand{\arraystretch}{1.8} 
% \begin{tabularx}{\textwidth}{|l|X|X|}
% \hline
% \textbf{Environment} & \textbf{Information Available} &  \textbf{Ambiguity Resolution } \\
% \hline
% \textbf{Base} & Both see both objects; Matcher near correct object; Director names it. & Ambiguity can be resolved  directly by processing initial demand. \\
% \hline
% \textbf{Persp} & Matcher sees both objects; Director sees only one. & Infers shared attention and selects mutually visible object. \\
% \hline
% \textbf{Near} & Both see both; Matcher near correct object. & Should confirm intention; avoid assumption from spatial closeness. \\
% \hline
% \textbf{Far} & Both see both objects; Matcher is near incorrect one. & Should not assume proximity = intent; seeks clarification. \\
% \hline
% \textbf{Hidden} & Object is visible to Director but hidden from Matcher. & Matcher must ask or explore, modeling unseen perspectives. \\
% \hline
% \textbf{NotThat} & Matcher sees only wrong object; Director sees both. & Must revise choice upon feedback; updates belief state. \\
% \hline
% \textbf{Distractor} & Matcher sees distractor; Director sees correct object elsewhere. & Matcher must trust correction and override own perception. \\
% \hline
% \end{tabularx}
% \vspace{0.8em}
% \caption{Environment Types and Their Perspective-Taking Challenges}
% \label{table_env}
% \end{table}
%
\begin{table}[h]
\centering
\small
\renewcommand{\arraystretch}{1.4}
\begin{tabularx}{\textwidth}{|p{2.35cm}|p{4cm}|p{3cm}|X|}
\hline
\textbf{Environment} & \textbf{Information State} & \textbf{Matcher Spatial State} & \textbf{Ambiguity Resolution } \\
\hline
\textbf{Base} & Both see both objects \& areas;  Director names  target explicitly.&
Distant from both \textit{target} and \textit{distractor}.
& Processing initial demand. \\
\hline
\textbf{Persp}ective Taking & Matcher sees both objects; Director sees only \textit{target} and one area less. &Close to both \textit{target} and  \textit{distractor}. & Using common ground.\\
\hline
\textbf{Dist}ractor & Matcher sees \textit{distractor}; Director sees \textit{target}. Each sees an area the other can’t. &Distant from \textit{target}, close to \textit{distractor}. &Exploring unseen perspectives. \\
\hline
\textbf{Near} & Both see both objects. Matcher sees an area more. & Close to \textit{target}, distant  from \textit{distractor}. & Asking clarification. \\
\hline
\textbf{Far} & Both see both objects. Matcher sees an area more.& Close to \textit{distractor}, distant from \textit{target}.& Asking clarification. \\
\hline
\textbf{Hidd}en & Director sees Target.  Matcher does not.  Each sees an area the other can’t. &Distant from \textit{target}. No Distractor.  &Exploring  unseen perspectives. \\
\hline
\textbf{Not} That & Matcher sees only \textit{distractor}; Director sees both.  Each sees an area the other can’t. &Close  to \textit{distractor}, distant from  \textit{target}.& Asking clarification. \\
\hline
\end{tabularx}
\vspace{0.8em}
\caption{Environment Types and Their Perspective-Taking Challenges}
\label{table_env}
\end{table}
In most cases, ambiguity is introduced by presenting two objects of the same type (e.g., two ties of different colours). This setup requires the Matcher to rely on cues beyond spatial proximity or direct visual recognition, such as dialogue with the Director or strategic exploration, to disambiguate the task. Figure  \ref{fig:env} shows such a design graphically.
To systematically vary perspective-taking difficulty, we designed seven task types, ranging from fully observable to highly ambiguous settings, as shown in Table \ref{table_env}.

The Matcher begins each trial at a random location and must perform a sequence of actions: moving, opening containers and optionally asking questions to infer the correct object. The agent must decide whether more information is required to make the choice and how to find it. This design enables controlled evaluation of perspective-taking under increasing complexity, balancing grounded environment interaction with higher-order inference demands.

\subsection{Strategy Generation}
\begin{figure}[ht]
    \centering
    \includegraphics[width=1\textwidth]{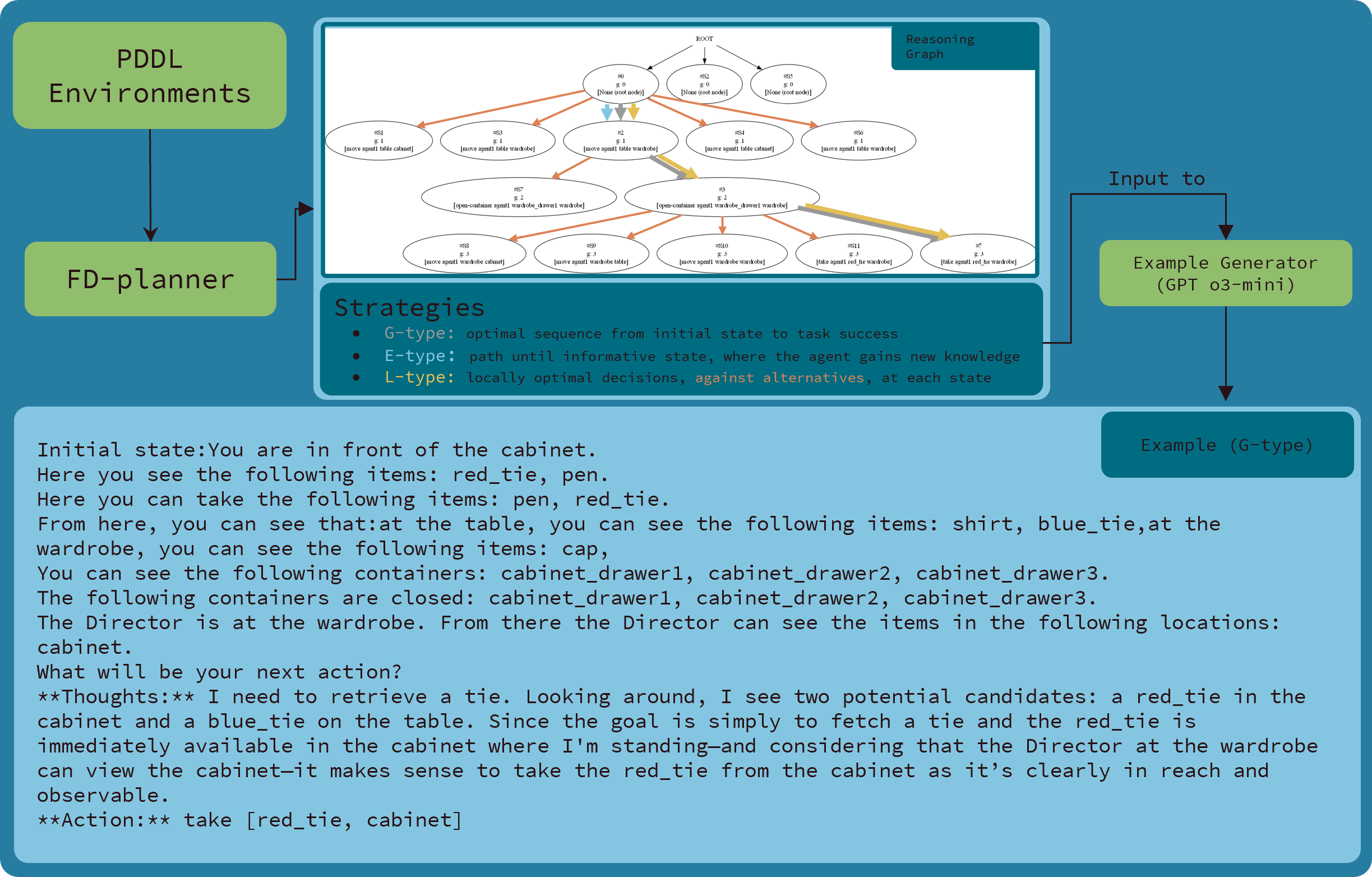}
    \caption{Graphic view of example generation pipeline. The panel above shows the reasoning tree produced by the Fast-Downward planner, highlighting the three different paths according to the corresponding strategy. The section below is an example (G-type path) generated from the Base environment. }
    \label{fig:ex_gen}
\end{figure}

% Add the 7 types of increasing persp taking tasks
To generate structured behavioral sequences, we employed a modified version of the Fast Downward planner, adapted to expose the internal reasoning process underlying its search, as shown in Figure \ref{fig:ex_gen}. Specifically, the planner was equipped to output a reasoning tree that records the sequence of states and actions considered during planning. This tree represents a detailed trace of the planner's decision space, capturing both explored and selected paths through the environment.
From this reasoning tree, we derived three types of training examples, each emphasizing a distinct mode of behavior, by extracting them through three different tree‐traversal strategies.:
\begin{itemize}
    \item Goal-directed trajectories (G-type): These are action sequences that lead from the initial state to the goal state, representing the planner's optimal solution path. They reflect efficient behavior under the assumption that the agent has access to all relevant information for completing the task.
    \item Information-seeking trajectories (E-type): These examples were extracted by identifying branches of the reasoning tree that led to states where critical information was gained, such as observing hidden objects or reducing ambiguity about the target. This type of example prioritizes epistemic actions and models behavior driven by uncertainty reduction rather than direct goal completion.
    \item Local decision points (L-type): For this example type, we identified the locally optimal action at each state visited during planning, contrasted explicitly with the other possible actions available at that point. This approach emphasizes granular decision-making and trains agents to reason about why the chosen action is superior to its alternatives given a single observation.
\end{itemize}
These examples are grounded in the planner's reasoning trace, providing insight into the deliberative process behind goal achievement and information gathering. While the Director is not explicitly modeled in the environment, its influence is indirectly embedded in the task configuration, for instance, through the target object’s proximity to the Director’s location, prompting the Matcher to infer spatial and referential cues. This setup supports the generation of cognitively rich example sequences from a formally grounded planning system.
\subsection{Example Generation}
The final step of the methodology involves transforming the planner-derived action sequences into structured cognitive examples that simulate the internal reasoning of an agent engaged in perspective-taking. These examples are constructed as ``thought-action'' pairs, where each action is preceded by a textual description of the agent's inferred reasoning process.

To produce these thought sequences, a large language model (GPT o3-mini) was prompted with each action sequence, drawn from the G-, E-, or L-type examples, along with contextual information such as the agent's current state, visible items, and the overall task objective. The model was asked to articulate the reasoning that might justify each step in the sequence, generating natural language explanations simulating deliberation and inference. The model was instructed to generate such sequences employing the following prompts:
\begin{table}[h]
\centering
{%
\renewcommand{\arraystretch}{1.4} % Only affects this table
\begin{tabular}{|l|p{10cm}|}
\hline
\textbf{Strategy} & \textbf{Prompt} \\
\hline
\textbf{G-type} & Given the sequence of actions the agent executed until reaching its goal in a specific scenario, reconstruct the agent’s reasoning step by step. Explain how each action contributed to achieving the goal. \\
\hline
\textbf{E-type} & Given a sequence of actions taken until the agent reaches an informative state (i.e., a state that provides new information), reconstruct the agent’s reasoning step by step. Describe how each action led to gaining information. \\
\hline
\textbf{L-type} & Given the agent’s last action, the set of possible actions, and the correct action in a specific scenario, explain the agent’s reasoning behind selecting that particular action over the alternatives. \\
\hline
\end{tabular}
}%
\vspace{0.8em}
\caption{Reasoning Strategy Prompts}
\label{tab:reasoning-strategies}
\end{table}

This transformation enriches the examples with cognitive structure, allowing them to reflect key faculties involved in perspective-taking. Each thought segment captures perceptual assessment (e.g., recognizing that an object is not currently visible), inferential reasoning (e.g., hypothesizing where a hidden item might be based on contextual cues), and decision-making under uncertainty (e.g., choosing to ask the Director when the situation is ambiguous). These elements are not explicitly modeled in the PDDL domain, but are essential for simulating realistic agent behavior in social contexts.

\section{Experiments}

We first augmented the Fast Downward planner ($A^{\star}$ with the admissible $h_{max}$ heuristic) to emit all evaluated actions during search, from which we built a complete reasoning tree. From this, we extracted three trajectory types, G-type, E-type, and L-type, each corresponding to a distinct decision-making strategy. These were fed to GPT o3-mini using strategy-specific prompts to generate natural-language thought–action chains: step-by-step goal reasoning (G-type), information-seeking justification (E-type), and local action selection rationale (L-type).

To evaluate the Matcher–Director interaction, we instantiated two LLM-agents (both GPT o3-mini) in our PDDL-based household environment. The stationary Director issued natural-language instructions of varying ambiguity; the ReAct-based Matcher alternated between LLM reasoning and grounded actions to interpret and execute the task. For each of the seven task types (ranging from disambiguated to highly ambiguous), the Matcher was trained on structured examples from the other six, and tested on the held-out one. Each test scenario was repeated across five trials. Performance was assessed using failure rate, step count, clarification queries, and epistemic actions.

In parallel, we enhanced the PDDL domain to allow the planner to compute optimal perspective-taking strategies by including epistemic (\texttt{ASK}) actions where ambiguity required them. This enabled the generation of example sets with (+ask) or without (–ask) clarification steps, providing expert baselines for reasoning under uncertainty and testing whether such exposure influenced agent behaviour.

To probe this further, we used these planner-optimal trajectories to generate an additional set of G-, E-, and L-type examples, enriched with epistemically aware reasoning but following the same prompting schema.

We report aggregated averages by task (see Tables \ref{tab:first-take-correct}, \ref{tab:steps}, \ref{tab:ask}) to compare how example types shape Matcher behaviour across perspective-taking demands. 
Despite encoding optimal epistemic strategies, these enriched examples did not yield consistent performance gains. Neither success rates nor behavioural quality (e.g., unnecessary queries, incorrect assumptions) improved significantly. Possible reasons for this outcome are explored in the following section.

\begin{table}[h]
\caption{First Take on Correct Target (\%) across Scenarios}
\label{tab:first-take-correct} %
\begin{tabularx}{\textwidth}{|p{3.2cm}|X|X|X|X|X|X|X|}
\hline
\textbf{Example type} & \textbf{Persp} & \textbf{Far} & \textbf{Hidd} & \textbf{Not} & \textbf{Dist} & \textbf{Base} & \textbf{Near} \\
\hline
G-type-ask      &  100 & 0 & 100 & 0 & 20 & 100 & 100\\
\hline
G-type+ask &  100 & 0& 100 & 0 & 40 & 100 & 100\\
\hline
E-type-ask      &  100 & 0 & 100 & 20 & 0 & 100 & 100  \\
\hline
E-type+ask&  100 & 0 & 100 & 0& 40 & 100 & 100  \\
\hline
L-type-ask      &  100 & 0 & 100 & 0 & 40 & 100 & 100 \\
\hline
L-type+ask &  100 & 0& 100 & 0 & 40 & 100 & 100 \\
\hline
No Examples &  100 & 20 & 100 & 0 & 40 & 100 & 100 \\
\hline
\end{tabularx}
\end{table}
\begin{table}[h]
\caption{Average Number of Steps}
\label{tab:steps}
\begin{tabularx}{\textwidth}{|p{3.2cm}|X|X|X|X|X|X|X|X|}
\hline
\textbf{Example type} & \textbf{Persp} & \textbf{Far} & \textbf{Hidd} & \textbf{Not} & \textbf{Dist} & \textbf{Base} & \textbf{Near} &\textbf{AVG}\\
\hline
G-type-ask &  1 & 4.8 & 3 & 5.8 & 5.2 & 2 & 1 & 3.26\\
\hline
G-type+ask &  1 & 5 & 3 & 5.8 & 5 & 2 & 1 & 3.26\\
\hline
E-type-ask &  1 & 4.4 & 3 & 6 & 5 & 2 & 1  & 3.2\\
\hline
E-type+ask &  1 & 4.6 & 3 & 5.6 & 5.2 & 2 & 1  & 3.2 \\
\hline
L-type-ask &  1 & 4.4 & 3 & 4.8 & 4.6 & 2 & 1 & \textbf{2.97}\\
\hline
L-type+ask &  1 & 5 & 3 & 5.8 & 5 & 2 & 1 & 3.26 \\
\hline
\textit{AVG} & \textit{1}	& \textit{4.7}	&\textit{ 3}	& \textit{5.63} &	\textit{5} &	\textit{2} &	\textit{1} &	\textit{3.19}\\
\hline
No Examples &  1 & 4.4 & 3 & 5.8 & 4.6 & 2 & 1 & 3.11\\
\hline
Planner &  1 & 3 & 2 & 3 & 2 & 2 & 2 & 2.14\\
\hline
\end{tabularx}
\end{table}
\begin{table}[h]
\caption{Average Number of \textit{Ask} Actions}
\label{tab:ask}
\begin{tabularx}{\textwidth}{|p{3.2cm}|X|X|X|X|X|X|X|X|}
\hline
\textbf{Example type} & \textbf{Persp} & \textbf{Far} & \textbf{Hidd} & \textbf{Not} & \textbf{Dist} & \textbf{Base} & \textbf{Near} & \textbf{AVG}\\
\hline
G-type-ask &  0 & 1.8 & 1 & 1.8 & 2.4 & 0 & 0 & 1\\
\hline
G-type+ask &  0 & 2 & 1 & 1.8 & 2.4 & 0 & 0 & 1.03\\
\hline
E-type-ask &  0 & 1.4 & 1 & 2.4 & 2 & 0 & 0  & 0.97\\
\hline
E-type+ask &  0 & 1.6 & 1 & 1.6 & 2.6 & 0 & 0  & 0.97\\
\hline
L-type-ask &  0 & 1.4 & 1 & 1.4 & 2 & 0 & 0 & \textbf{0.83}\\
\hline
L-type+ask &  0 & 2 & 1 & 1.8 & 2.6 & 0 & 0 & 1.06\\
\hline
\textit{AVG} & \textit{0}	& \textit{1.7} &	\textit{1} &	\textit{1.8} &	\textit{2.33} &	\textit{0}	& \textit{0} &	\textit{0.98}\\
\hline
No Examples &  0 & 1.6 & 1 & 1.8 & 2 & 0 & 0 & 0.91 \\
\hline
Planner & 0 & 1 & 0 & 1 & 0 & 0 & 1  & 0.43\\
\hline
\end{tabularx}
\end{table}

\section{Discussion}\label{discussion}
Our results reveal that success in collaborative reference hinges on
\emph{three orthogonal but interacting cognitive demands}.  Throughout we denote them as \textbf{F1-F3} to emphasise their functional independence.
\begin{description}
\item[F1 Common-ground filtering.]  Listeners must inhibit any object that the Director cannot currently see, a Level-1 perspective-taking operation that relies mainly on selective attention \cite{keysar1997unconfounding,keysar2003limits}.
\item[F2  Imagining Director-privileged space.]  When the layout contains occluded regions visible only to the Director, the agent must
\emph{construct counterfactual scenes} and reason about what may be present there, a genuinely mentalistic computation that recruits full Theory of Mind (ToM) \cite{bianco2019functional}.
\item[F3  Metacognitive cost-benefit evaluation.]  Because belief tracking, exploration, and clarifying questions all incur different costs, the agent must decide whether to pay those costs or rely on faster egocentric heuristics \cite{patania2024large,tversky1974judgment,russell1991principles,oliehoek2016concise,griffiths2019doing}.
\end{description}

\paragraph{When do the factors matter?}
Whenever the task could be solved by \textbf{F1} alone, as in \textit{Base}, \textit{Perspective-Taking}, and \textit{Near} first-take accuracy was perfect.  Introducing \textbf{F2} precipitated large drops (\textit{Distractor}, \textit{Hidden}, \textit{Not That}); accuracy rebounded only after the agent executed an exploratory `look-inside' or issued a query.  Pure \textbf{F3} pressure (\textit{Far}) likewise reduced first-take accuracy, even though no ToM inference was required.  Table~\ref{tab:first-take-correct} quantifies this triple dissociation.

\paragraph{Selective attention versus Theory of Mind.}
The split between \textbf{F1} and \textbf{F2} mirrors the neuroscientific dissociation between the dorsal rTPJ region, associated with social ToM reasoning, and a neighbouring ventral patch specialised for attentional re-orienting \cite{igelstrom2016topographical,bitsch2018role}.  
Our LLM agent performs the attentional filter (\textbf{F1}) flawlessly but fails when genuine belief reasoning (\textbf{F2}) is needed, reproducing the human pattern reported by \cite{keysar2003limits}.

\paragraph{Why do planner-optimal traces fail to transfer?}
Fast-Downward plans embody the invariant `\emph{act only once the target is uniquely identifiable to the Director}'.  Yet few-shot exposure to those traces did not boost performance because (i) GPT-o3 implicitly assigns near-zero cost to questions, so it still `plays safe,' and (ii) the linear action lists do not make the underlying cost rationale explicit.  Without an explicit belief state and cost model, the LLM reverts to the heuristic `grab the closest match, ask if unsure'.

\paragraph{Prompting design for exploration}
Beyond cost modelling, an additional limitation concerns the formulation of prompts used to frame the problem and describe the scenario. Several failures related to \textbf{F2} and \textbf{F3} occurred in situations where the agent would need to hypothesise the existence of relevant but currently unseen objects, an ability central to active vision. These may result from prompts that insufficiently foreground the plausibility of missing information in unexplored, yet accessible, regions. Without an explicit representation of such areas, the agent may not infer that exploration is necessary or worthwhile. Prompt strategies that better support uncertainty reasoning and hypothesis generation about occluded content could thus be essential for more robust epistemic behaviour.

\paragraph{Limitations and future directions.}
The present grid world offers only binary visibility; richer social environments include graded salience, gaze cues, and competing conversational goals.  Extending the benchmark to those dimensions should exert stronger pressure on \textbf{F2} and \textbf{F3}, revealing whether the same failure modes persist or whether multi-modal grounding and cost signals can scaffold more robust ToM.

\section{Conclusions}
%We find that a ReAct loop backed by GPT-o3 already supports flawless Level-1 perspective taking (\textbf{F1}), but remains brittle on two fronts: (i) \textbf{F2}, imagining alternative objects in occluded space, and (ii) \textbf{F3}, weighing the immediate versus delayed costs of belief-driven action.  Bridging these gaps may require \emph{explicit belief state tracking} \cite{bianco2022robot,ognibene2013contextual}, \emph{learned cost models}, and \emph{prompting strategies that foreground hidden regions}.  Testing agents in richer, more uncertain settings extracted from robot sensors, and penalising gratuitous queries, will be essential for advancing from attentional filtering to full, cost-aware Theory of Mind competence.
We find that a ReAct loop backed by GPT-o3 already supports flawless Level-1 perspective taking (\textbf{F1}), but remains brittle on two fronts: (i) \textbf{F2}, imagining alternative objects in occluded space, and (ii) \textbf{F3}, weighing the immediate versus delayed costs of belief-driven action. Bridging these gaps may require \emph{explicit belief state tracking} \cite{bianco2022robot,ognibene2013contextual}, \emph{learned cost models}, and \emph{prompting strategies that foreground hidden regions and explicitly stimulate hypothesis generation about unseen content}. Testing agents in richer, more uncertain settings extracted from robot sensors, and penalising gratuitous queries, will be essential for advancing from attentional filtering to full, cost-aware Theory of Mind competence.

\subsubsection{\ackname}
This work was supported by the Volkswagen Foundation under the funding programme “Open Up – New Research Spaces for the Humanities and Cultural Studies,” project “Developing an Artificial Social Childhood (ASC) to improve AI causal reasoning, information gathering and decision making,” reference 9E530.

\bibliographystyle{splncs04}
\bibliography{mybib}
\end{document}